\documentclass[letterpaper,twocolumn,fleqn]{article} 

\usepackage{ist}
\usepackage{subfigure}
\usepackage{url}

\title{A Visual Technique to Analyze Flow of Information in a Machine Learning System}

\author{ Abon Chaudhuri, Walmart Labs, Sunnyvale, CA, USA}

\date{} % date has an empty field.

\begin{document} 

\maketitle 

\thispagestyle{empty} % prevents the first page to be numbered

%%%%%%%%%%%%%%%%%%%%%%%%%%%%%%%%%%
% Abstract
%%%%%%%%%%%%%%%%%%%%%%%%%%%%%%%%%%

\begin{abstract}
Machine learning (ML) algorithms and machine learning based software systems implicitly or explicitly involve complex flow of information between various entities such as training data, feature space, validation set and results. Understanding the statistical distribution of such information and how they flow from one entity to another influence the operation and correctness of such systems, especially in large-scale applications that perform classification or prediction in real time. In this paper, we propose a visual approach to understand and analyze flow of information during model training and serving phases. We build the visualizations using a technique called Sankey Diagram - conventionally used to understand data flow among sets - to address various use cases of in a machine learning system. We demonstrate how the proposed technique, tweaked and twisted to suit a classification problem, can play a critical role in better understanding of the training data, the features, and the classifier performance. We also discuss how this technique enables diagnostic analysis of model predictions and comparative analysis of predictions from multiple classifiers. The proposed concept is illustrated with the example of categorization of millions of products in the e-commerce domain - a multi-class hierarchical classification problem.  
\end{abstract}

\section{Introduction}
\label{sec:intro}
Machine learning (ML) algorithms and machine learning based software systems implicitly or explicitly involve complex flow of information between various entities such as training data, feature space, validation set and results. Each of them contains valuable information. The quality, quantity, and distribution of such information across different containers, and their flow from one to another influence the operation and the correctness of machine learning based systems. Certain algorithms such as probabilistic graphical models or deep neural networks explicitly rely on the flow of information. Employing statistical and visual methods to understand the distributions and the flow of information is critical to the success of large-scale data science applications. 

Let us consider a real application - product categorization - a large-scale classification task commonly encountered in the e-commerce domain (for classifying commodities into thousands of categories) for example.  The problem is to assign every product a category from a multi-level hierarchy of categories such as ``home$\rightarrow$kitchen$\rightarrow$appliances$\rightarrow$microwave".  To develop an accurate model (a classifier) for such a task, it is crucial to answer a number of key questions at every step - starting from data collection to feature engineering, training, and finally, evaluation. A few examples are: is every class well represented in the training data? Are there redundant features or collinearity among features? Does the evaluation strategy cover examples from all the classes? In addition to statistical analysis, the use of visual analytics to answer these questions effectively is becoming increasingly popular.  

Going one step deeper, we observe that the flow of information across various entities can often be formulated as joint or conditional probability distributions. A few examples are: distribution of class labels in the training data, conditional distribution feature values given a label, comparison between distribution of classes in test and training data. Statistical measures such as mean and variance have well-known limitations in understanding distributions. On the other hand, visualization based techniques allow a human expert to analyze information at different levels of granularity. To give a simple example, a histogram can be used to examine different sub-ranges of a probability distribution. In this paper, we present how Sankey diagrams~\footnote{https://developers.google.com/chart/interactive/docs/gallery/sankey} can represent probability distributions at various levels of detail. While this is not a new technique, we reinvent it as a visual encoding for joint and conditional distributions. Use of this particular technique along with supporting visualization techniques can lead to effective visualization-enhanced machine learning systems.

We also discuss the use of this technique in model comparison and diagnostics. For a given task, multiple models with different features and parameters are usually trained at the same time or over a period of time. When it comes to select the best one for large-scale use, the proposed technique allows a human expert to study relevant questions such as if they were trained on near identical data, if their performance varied significantly across certain categories - as opposed to relying only on the overall accuracy number.

\section{Related Work}
\label{sec_rel_work}
\paragraph{Visualization in Machine Learning:} Recently, both machine learning and visualization research communities have started to adopt techniques from each other, leading to publications and open source software systems for visually exploring different stages of a machine learning pipeline. 

Alsallakh et al.~\cite{visual1} propose to use multiple box plots to visualize the feature distribution of training samples and the ability of a feature to separate data into different classes. FeatureInsight~\cite{featureinsight} is a system that combines human and machine intelligence to examine classification errors to identify and build potentially useful features. Infuse~\cite{infuse} is another visual analytic system that allows the human analyst to visually compare and rank different features based on their usefulness in model building.

Another area of active research is understanding of machine learning and deep learning algorithms. Visual methods have been proposed to construct and understand classifiers using traditional machine learning algorithms such as Bayesian modeling~\cite{bayes_vis}, Decision Tree~\cite{dctree_vis} and Support Vector Machine~\cite{svm_vis}. Of late, visualization has turned out to be a candidate approach to understand deep learning models which are inherently harder to explain. A growing body of work~\cite{nodelink_cnn, hidden, activis} explore different ways to understand the evolution of feature maps, activations, and gradients in deep neural nets. t-SNE~\cite{tsne2008} a powerful technique for visual exploration of high-dimensional embeddings of data often produced by deep learning algorithms.

A number of recent works explore visual techniques to interpret the results of a machine learning algorithm. Augmented confusion matrices and confusion wheels~\cite{visual1} can effectively highlight instances or classes that a classifier is more likely to classify incorrectly. Modeltracker~\cite{modeltracker} presents an enhanced visual error analysis tool that complements and enhances the traditional measures for model performance.

Chen et al.~\cite{vis_survey} conducts a design study of the usefulness of a number of visual techniques used in machine learning and presents a system called VizML that allows diagnostic analysis of machine learning models. 

Besides academic research articles, resources such as tutorials and blogs~\cite{mlvisblog} summarizing the visualization techniques commonly used in machine learning are widely available. 
\paragraph{Visualization of Data Distributions:} Understanding the data distribution often holds the key to statistical modeling. Visualization is often more descriptive compared to summary statistics such as central and higher order moments. Histograms and box plots continue to be the most popular techniques to visualize discretized probability distributions. Violin plots have been used in certain cases. They are the popular choices when building more complex visualization software involving distribution data. Potter et al.~\cite{distvis2} presents a comprehensive summary of the techniques used to visualize probability density functions and cumulative distribution functions.

\section{Proposed Visualization Methodology}
\label{sec_main_technique}
A machine learning algorithm tries to learn the parameters of a mathematical model from known examples of a dataset, and uses the same model to predict information about unknown examples of that dataset. Hence, maximizing the use of the information present in the known examples is critical for successful modeling. In this paper, we propose to consider the input data, the intermediate transformations of the data, and the output (the predictions) as sets containing information.

The training data contains a set of instances $X$ and a set of labels $Y$. Usually, all or some of the instances have labels. In a standard supervised classification task, many instances map to a label.

The training instances are often converted to features using some mathematical function. We denote the features obtained by applying a function $f$ on $X$ by $F=f(x)$. Usually, a large number of features is computed from the data. In the presence of $m$ features, $F_k=f_k(x)$ denotes the $k^{th}$ feature dimension where $k$ varies from 1 to $m$.

After training an algorithm, it is usually evaluated on a held-out set or evaluation set of instances. We denote that set as $R$. The true labels of these instances (denoted by $Y_g$) are already known, the predictions or the actual labels produced by the algorithm (denoted by $Y_p$) is compared against these ground truth information.

Table~\ref{tbl:common_sets} lists the above mentioned entities or information containers. Technically speaking, we present them as multi-sets or bags so that the duplicates (example: the training labels repeat many times in a dataset). Also, given that the training and evaluation is inherently an iterative process, we add the notion of time or instance using a superscripted suffix $t$. Depending on the context, this suffix may denote a particular date or simply the iteration number.
\begin {table}[tb]
\caption {Common Notations to Represent Information Containers in a Machine Learning System}
\label{tbl:common_sets}
\begin{center}
\begin{tabular}{ |c|c|c| } 
 \hline
 Set & Notation & Size    \\ \hline
 Training Set & $X^t$    & $N^t$   \\ 
 Training Set Labels & $Y^t$ & $N^t$   \\ 
 $k^{th}$ Feature & $F_k^t$ & $N^t$   \\ 
 Evaluation Set & $R^t$ & $N_E^t$ \\ 
 Evaluation Set Labels & $Y_g^t$  & $N_R^t$ \\
 Evaluation Set Predictions & $Y_p^t$  & $N_E^t$ \\ \hline
 \end{tabular}
\end{center}
\end{table}

In a machine learning system, the relationships between such sets or multi-sets are often captured by conditional or joint probability distributions. We propose to visually capture these relationships using a technique called  Sankey diagrams~\footnote{http://www.sankey-diagrams.com} that was traditionally used to visualize flow from one set to another. We re-purpose this technique to meaningfully visualize flow of information some of which can be formulated as joint and conditional probability distributions. 

\begin{figure}[tb]
    \centering
     \includegraphics[width=0.75\linewidth]{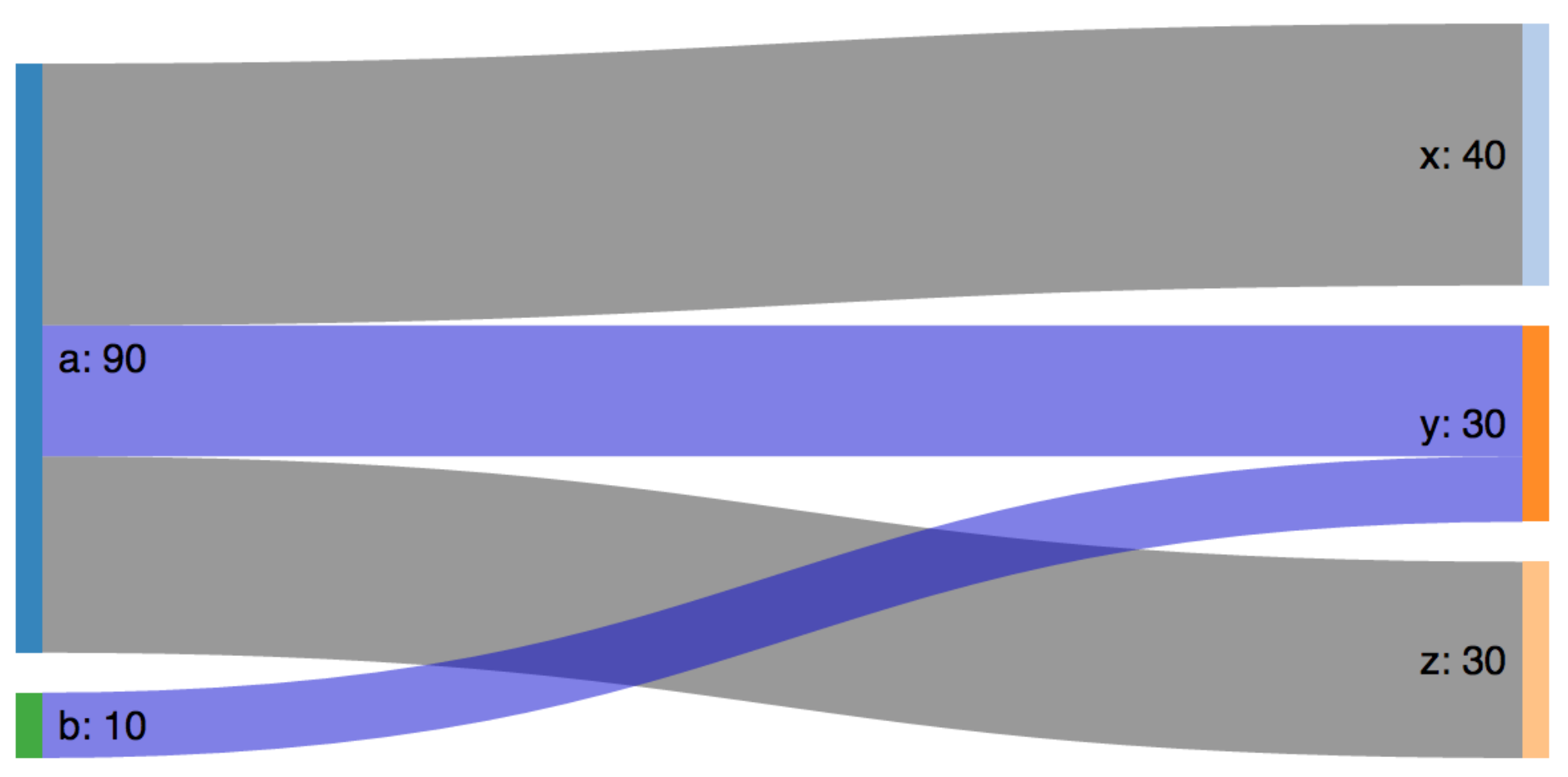}
     \caption{Sankey diagram as a visual expression of a conditional probability distribution. In this example, the purple lines highlight $P(T=y|S=a)$ (value=0.2) and $P(T=y|S=b)$ (value=1.0)}
     \label{fig:sankey_basic}
\end{figure}

Let us consider an example with two multi-sets, each containing 100 elements - $S = \{a: 90, b:10\}$ and $T = \{x:40, y:30, z:40\}$. In other words, $S$ is a bag of 90 instances of $a$ and 10 instances of $b$ and so on. Suppose, we are interested in the conditional distribution $P(T|S)$. More specifically, we would like to compare $P(T=y|S=a)$ (value=0.2) with $P(T=y|S=b)$ (value=1.0). The probability values hardly explain anything about the data. On the contrary, the Sankey diagram in Figure~\ref{fig:sankey_basic} (follow the two purple flows) reveals much more about the data than the the probabilities. For example, it clearly shows that despite the higher value of $P(T=y|S=b)$, it includes only a narrow subset of elements since $P(S=b)$ is quite low. Rest of paper will present examples of such insights using Sankey diagrams.
\subsection{Application Scenario}
\label{sec_application}
\begin{figure}[tb]
    \centering
     \includegraphics[width=0.98\linewidth]{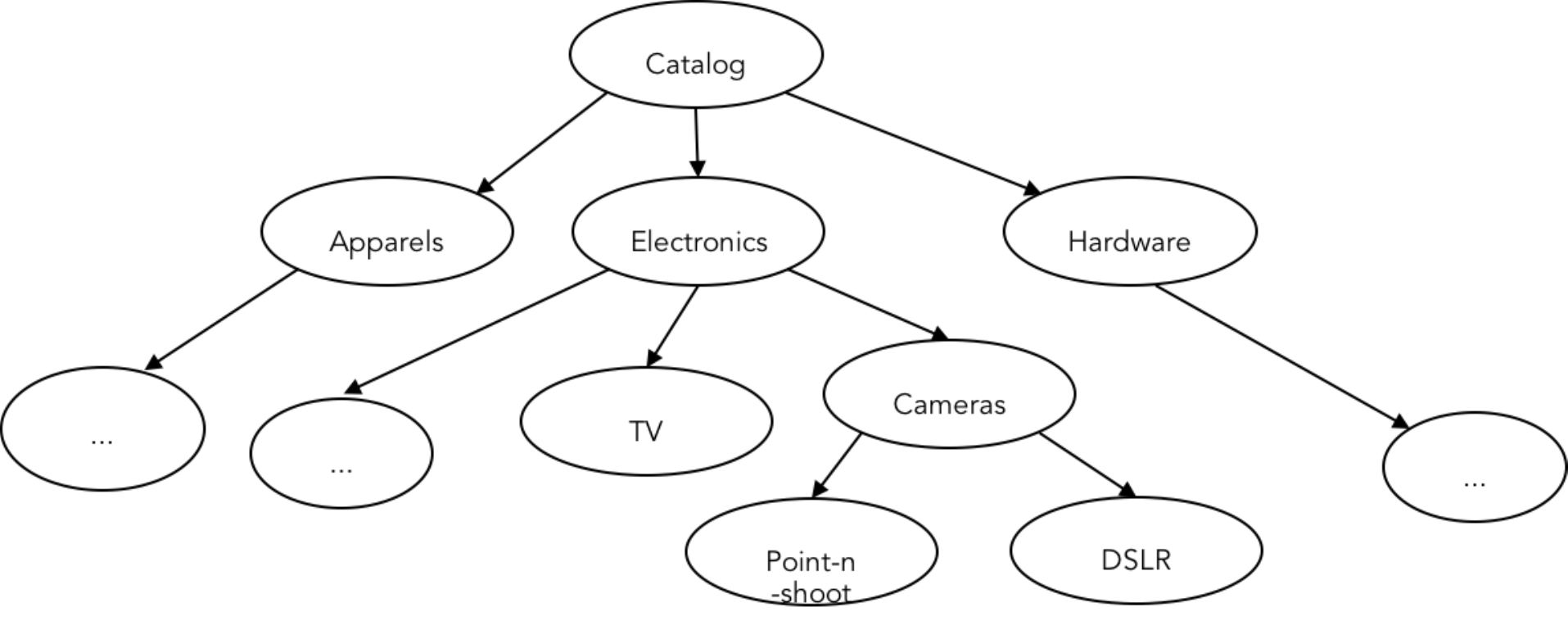}
     \caption{A simplified illustration of a multi-level product category hierarchy.}
     \label{fig_taxonomy}
\end{figure}

The product catalog of an e-commerce company typically contains millions of products that need to be placed into categories structured as multi-level hierarchies (such as electronics$\rightarrow$appliances$\rightarrow$kitchen appliances$\rightarrow$microwave). An oversimplified example is shown in Figure~\ref{fig_taxonomy}. Categorization ensures that the product is visible on the correct segment of the website and searchable by the users. 

From a machine learning point of view, this is a classic multi-class multi-label hierarchical classification problem. Typically, a training data of a few million products is used to train four to five level hierarchy of categories containing a few thousand leaf level categories. Product name, description, image and a number of other properties are leveraged to compute features for the model~\cite{wmt_classification}. The trained model is evaluated using a controlled and sampled set of at least a few thousands products. 

This application can benefit heavily from visual exploration of data because the scale and complexity of the data is enormous, the category hierarchy is deep and ever-changing. Finally, visual exploration leads to data cleaning or filtering actions, causing incremental yet quick improvements. The following sections uses this as the primary case study.
\subsection{Visual Analysis of Training Data}
\label{subsec_training_data}
The training data consists of items and labels. In the context of product categorization, each instance of the training data contains product information such as title and other attributes, and category name which serves as the class label. Understanding the distribution of these labels in terms of quantity and quality is a key to building an accurate classifier. 
\subsubsection{Distribution of Relative Label Quantity}
\label{subsec_trdata_dist} 
The training dataset should have adequate examples from each class for the algorithm to learn. Given a hierarchical training dataset, the quantity of data should be adequate at each level of the hierarchy. Also, the distribution of labeled data among the children (classes) of each class should be nearly uniform. Let us assume a class $C_l$ has three children $C^1_{l+1}$, $C^2_{l+1}$, and $C^3_{l+1}$. Even if $C_l$ has enough training instances, if most of it comes from one of the child classes, the model may become biased towards that class and tend to mis-classify the other two classes to that one. 

To give an example from our application, it is not enough to have sufficient training examples for a category called "Camera". If "Camera" has two finer sub-categories "DSLR" and "Point-n-shoot", the total number of labels should be near equally distributed between these two. Otherwise, the classifier will tend to classify most cameras to the dominant sub-category. 

\begin{figure}[tb]
    \centering
   \begin{subfigure}[Nearly uniform distribution of labels among child classes of "Baby"]
       {\label{fig_tr_balanced}\includegraphics[width=0.45\linewidth]{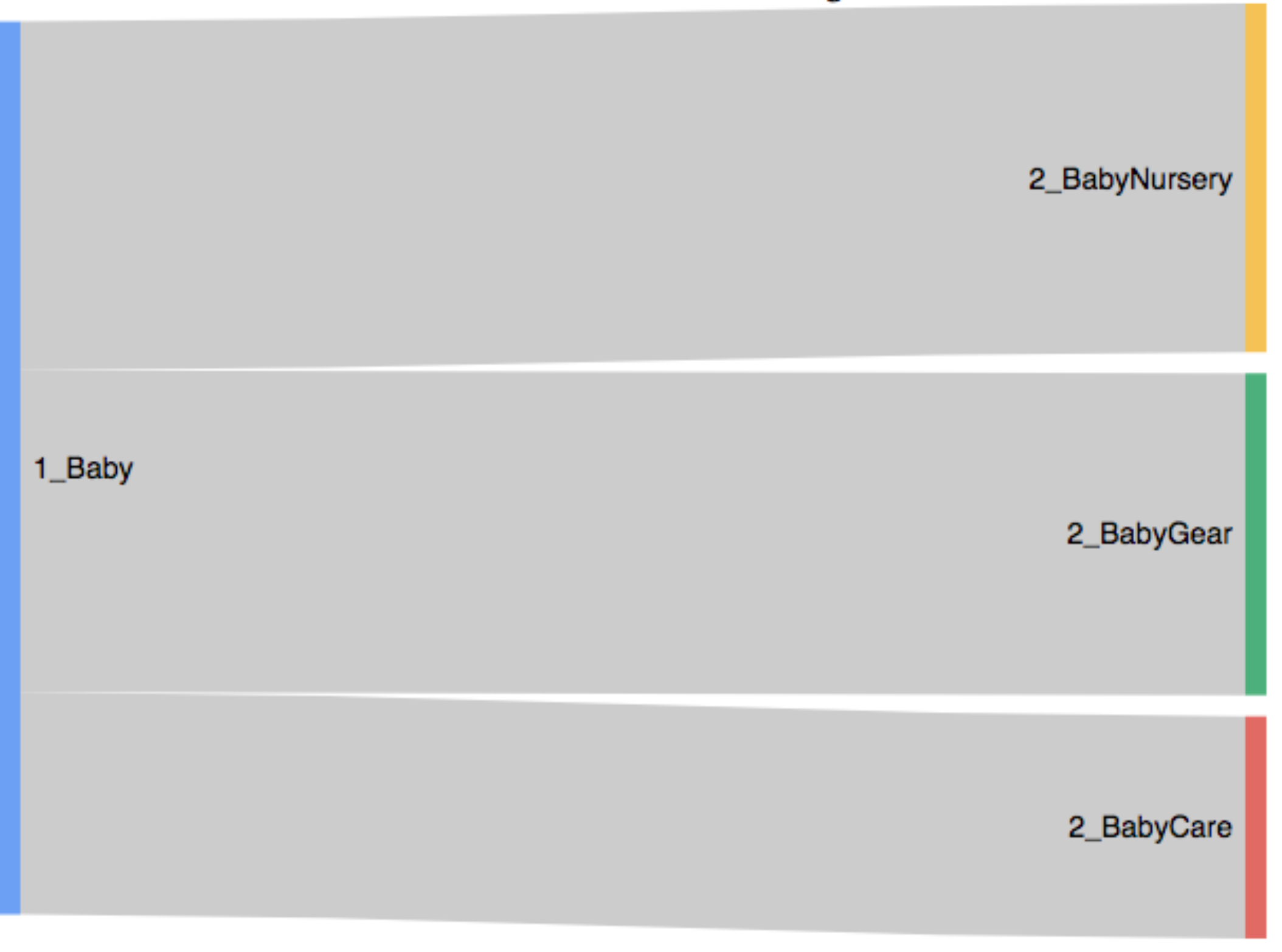}}
    \end{subfigure}
    \begin{subfigure}[Disparity among the relative quantity of labels among the child classes of "Junior"]
  {\label{fig_tr_unbalanced}\includegraphics[width=0.45\linewidth]{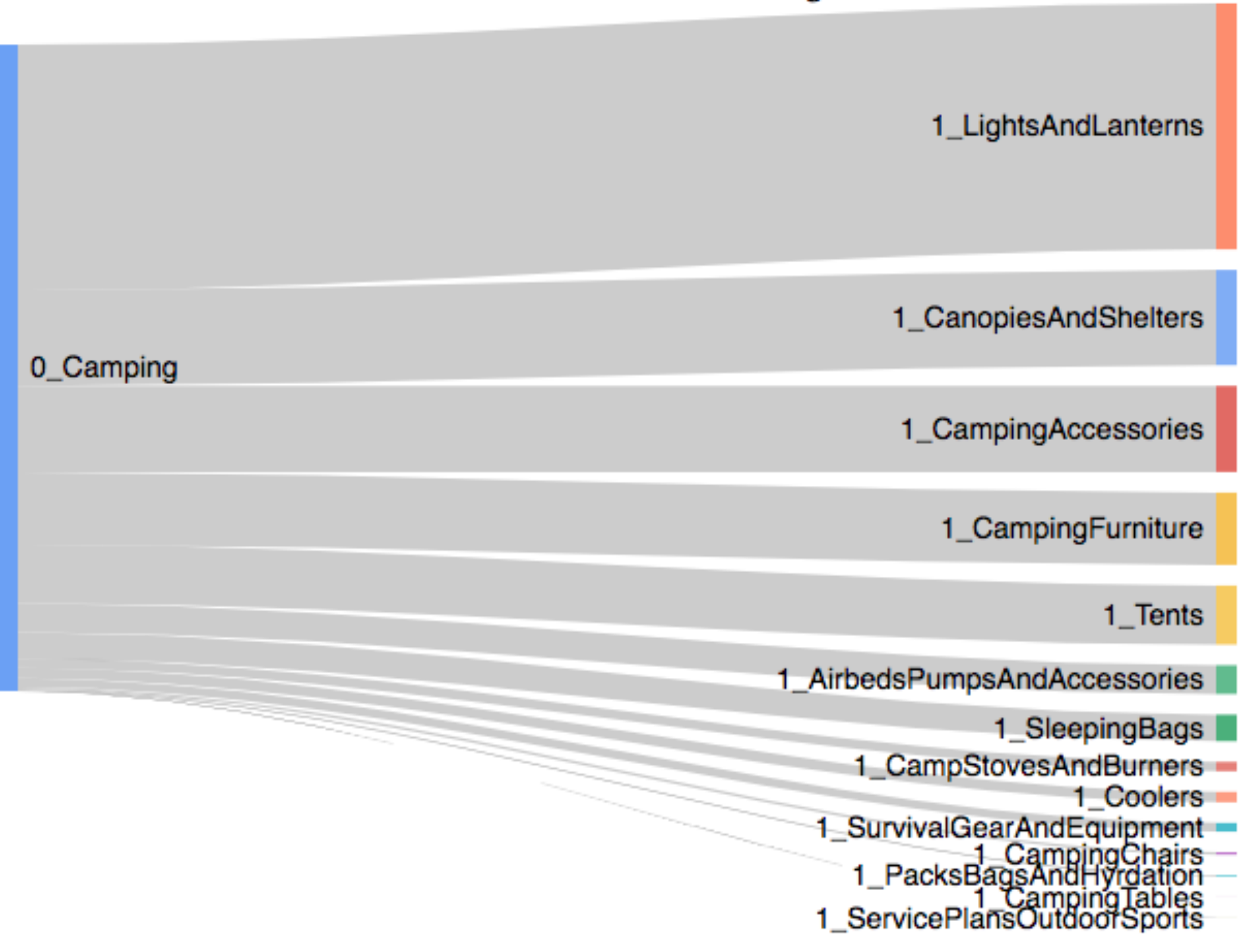}}
   \end{subfigure}
   \caption{Use of Sankey diagram to visualize the relative distribution of training labels among finer sub-categories}
    \label{fig_tr_label_dist}
\end{figure}

The absolute quantity of labels at each node can be visualized in many ways. To capture the relative distribution among the child classes, we propose to use Sankey diagrams. One possible implementation is to determine the width of the Sankey diagram's $i^{th}$ right side node based on $\frac{N_i}{\sum{N_k}}$ where $k$ iterates over all the child nodes in consideration. Figure~\ref{fig_tr_balanced} presents a near uniform label distribution among the child classes under "Baby". On the contrary, Figure~\ref{fig_tr_unbalanced} shows an example where certain classes under "Camping" do not have enough training examples. 

\begin{figure}[tb]
	 \includegraphics[width=\linewidth]{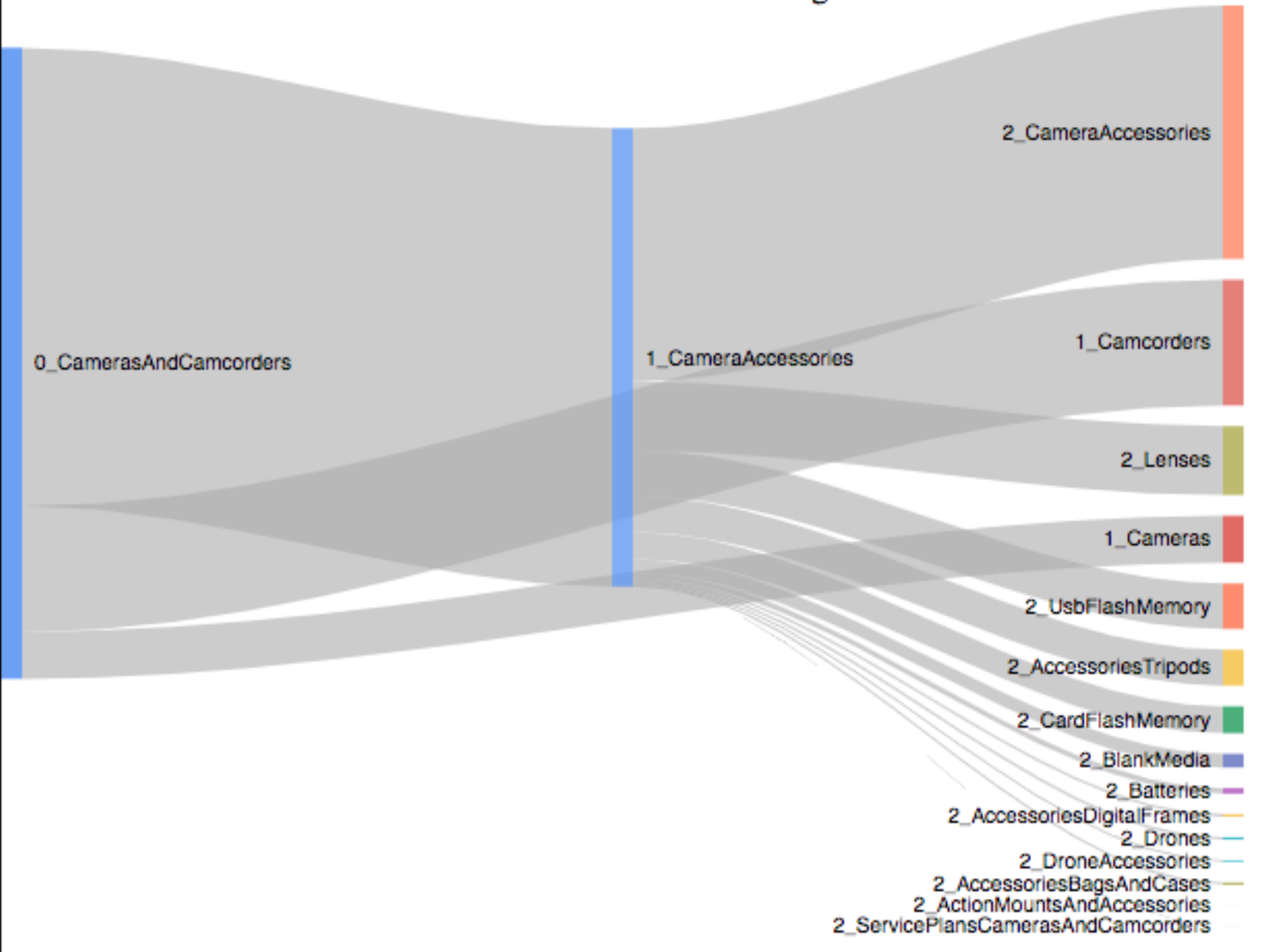}
      \caption{Use of multi-level Sankey diagram to show the distribution of labels from top to level of the class hierarchy. "Tripod" and "Flash memory" is sparsely populated categories compared to "Camcorders" and "Lenses". The prefix of the label indicates the level at which the category is located in the hierarchy. For example, "Cameras" is at the intermediate level where "Lenses" is at the leaf level.}
     \label{fig_sankey_trflow}
\end{figure}

We extend this idea to visually inspect an entire path of the hierarchy using multi-level flow diagrams (Figure ~\ref{fig_sankey_trflow}). This is reveal imbalance at inner levels. For example, even though "DSLR" and "Point-n-shoot" both have sufficient training data, "Cameras" itself may have low relative quantity compared to other electronics categories.
\begin{figure}[tb]
   \centering
   \begin{subfigure}[A category called "Windows" with well-balanced distribution of label quality.] 		    {\label{fig:labeldist_1}\includegraphics[width=0.45\linewidth]{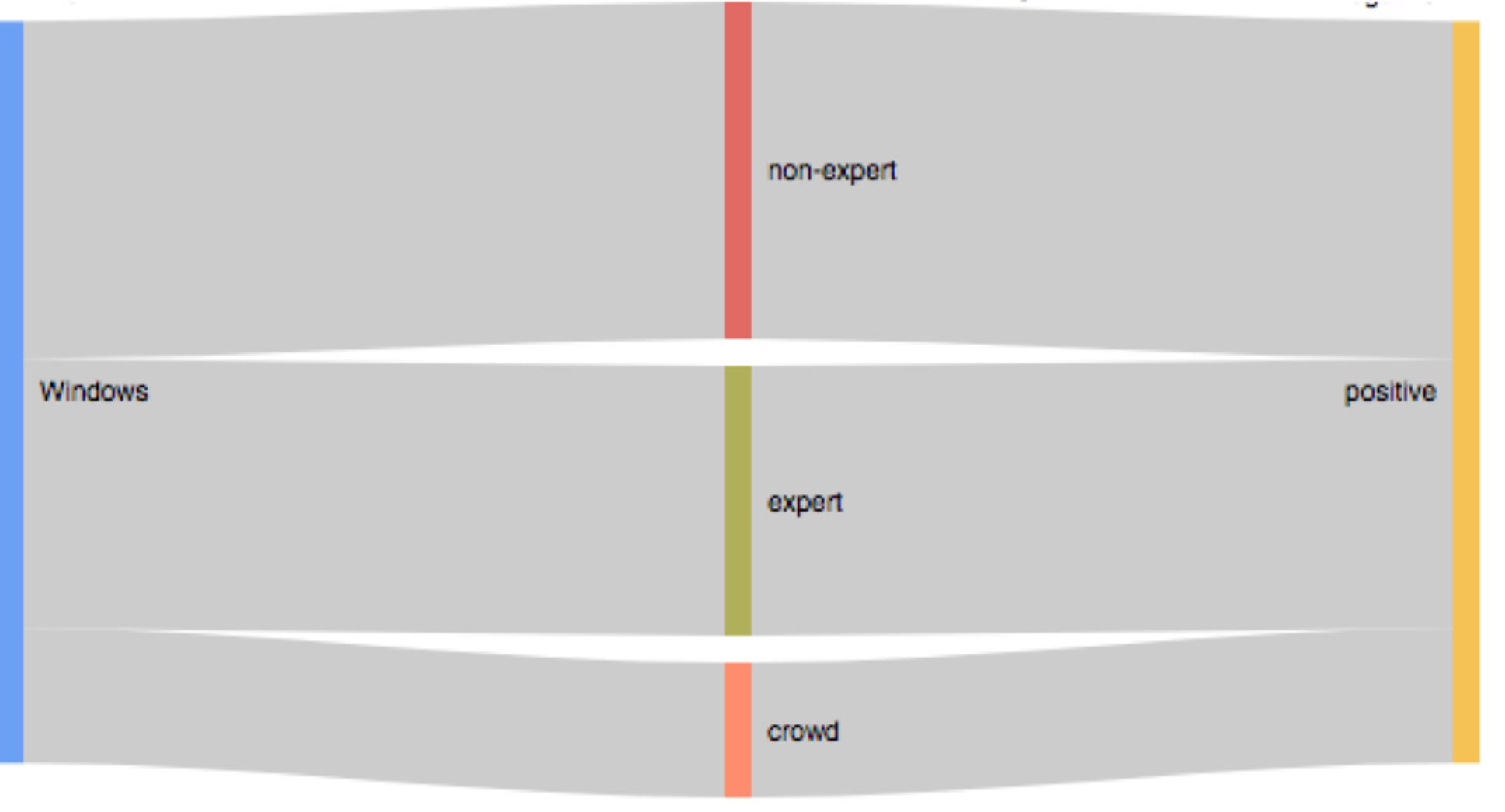}}
    \end{subfigure}
    \begin{subfigure}[A category called "Traditional Camcorders" with majority of low quality labels and some negative labels]
  {\label{fig:labeldist_2}\includegraphics[width=0.45\linewidth]{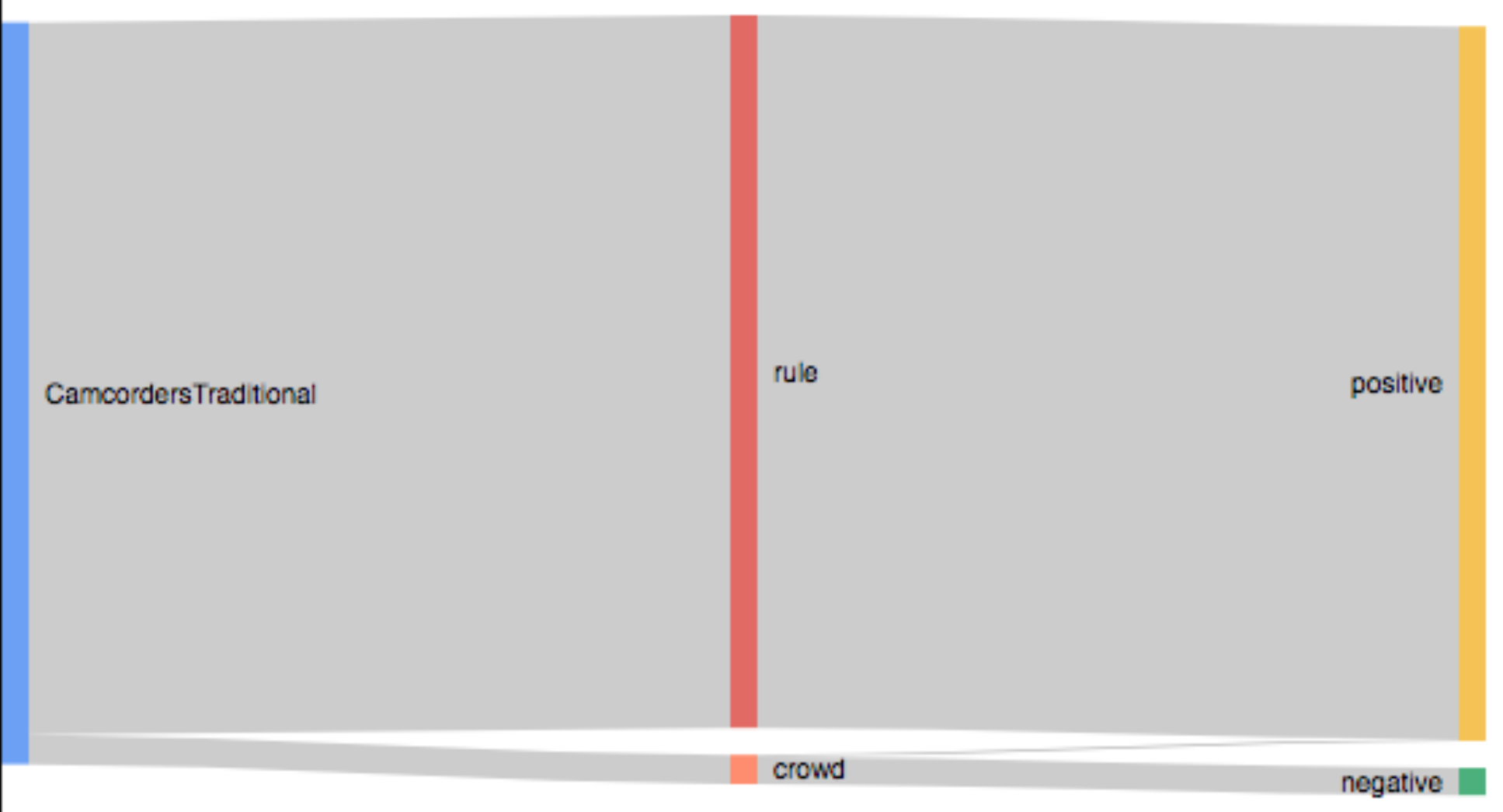}}
   \end{subfigure}
   \caption{Use of Sankey diagram to visualize the distribution of labelS collected from different sources and the distribution of positive and negative labels.}
    \label{fig_tr_label_dist}
\end{figure}
\subsubsection{Distribution of Label Quality}
\label{subsec_trdata_dist} 
Training instances in real applications are not curated. They are often noisy and incorrect, demanding quality assessment and control before they are used to train an algorithm. They often come from variety of sources marked with accuracy levels, making the inspection of quality relatively easier. For example, crowd-sourced labels are usually less trusted than labels annotated by trained experts. Also, labels may become outdated over time. In case of our e-commerce application, this happens because products are continuously incorporated into and taken off the catalog. 

It is common to have a training dataset with both positive and negative examples. For our product categorization example, the dataset often has labels that indicate that a product does not fall into a category. For example, "this item is not a DSLR camera". These type of labels are usually generated by curating the predictions of an existing model by human experts. If the training algorithm cannot leverage negative labels, then distribution biased towards positive labels is welcome. Otherwise, a balanced distribution is favorable.

We inspect the quality of training labels at least from two aspects: decision type (positive or negative) and source. For each category, we use Sankey diagram to inspect the distribution of labels from various sources and for each source, various decisions. The goal here is two-fold:
\begin{itemize}
\item to identify the classes for which the distribution is well-balanced: High accuracy is expected off these classes. If not, further investigation is recommended post evaluation. An example is shown in Figure~\ref{fig:labeldist_1}.
\item to identify the classes for which the distribution has high proportion of less reliable source(s). These classes require immediate attention. An example is shown in Figure~\ref{fig:labeldist_2}.
\end{itemize}
\subsection{Visual Analysis of Features}
\label{subsec_feature}
In a typical modeling problem, hundreds, if not thousands, of features are extracted from the training dataset. To prevent over-fitting of the model and to make the best use of computation power and storage, it is important to train only on features that are important. However, the notion of importance is not well-defined. Some statistical measures such as Welch's statistic can be used to see if a feature has the ability to separate data into classes. We propose that visual methods provide certain types of insight that cannot be derived from statistical methods.
\subsubsection{Feature Importance Estimation}
\label{sec_feature_imp}
\begin{figure}[tb]
	 \includegraphics[width=\linewidth]{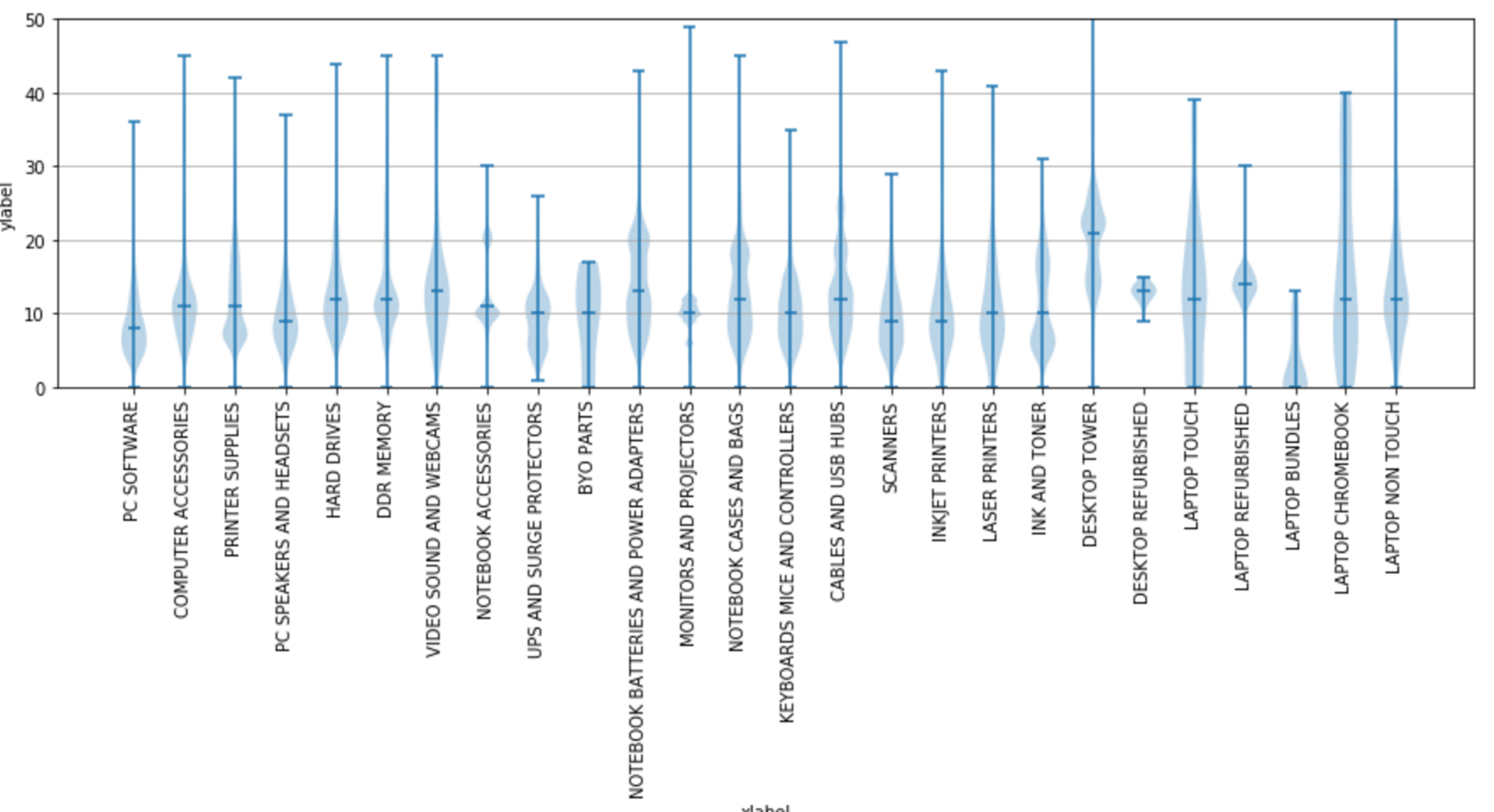}
      \caption{Comparative visualization of the distributions of lengths of product titles from a few semantically close classes. For example, \textit{Notebook batteries} and \textit{Notebook cases} have almost identical distributions. The title length is not a good feature between these two classes.}
     \label{fig:trdata_violin}
\end{figure}

\textbf{Numeric Features:} In general, a feature can help in classifying data if its presence is distinctive in different classes. For example, if the range of a numeric feature over class $A$ do not overlap with the range of the same feature values for class $B$, this feature can help differentiate the two classes. Hence, visualizing the distributions of a numeric feature for different categories can be immensely helpful. Our product categorization example relies heavily on product titles. Length of product title is an important numeric feature. We propose to examine the distribution of the length of product titles for different categories using violin plots (Figure~\ref{fig:trdata_violin}).

\begin{figure}[h]
    \centering
    \begin{subfigure}[Sankey diagram of a categorical feature less relevant to a classification task]
       {\label{fig:sankey_nr}\includegraphics[width=0.45\linewidth]{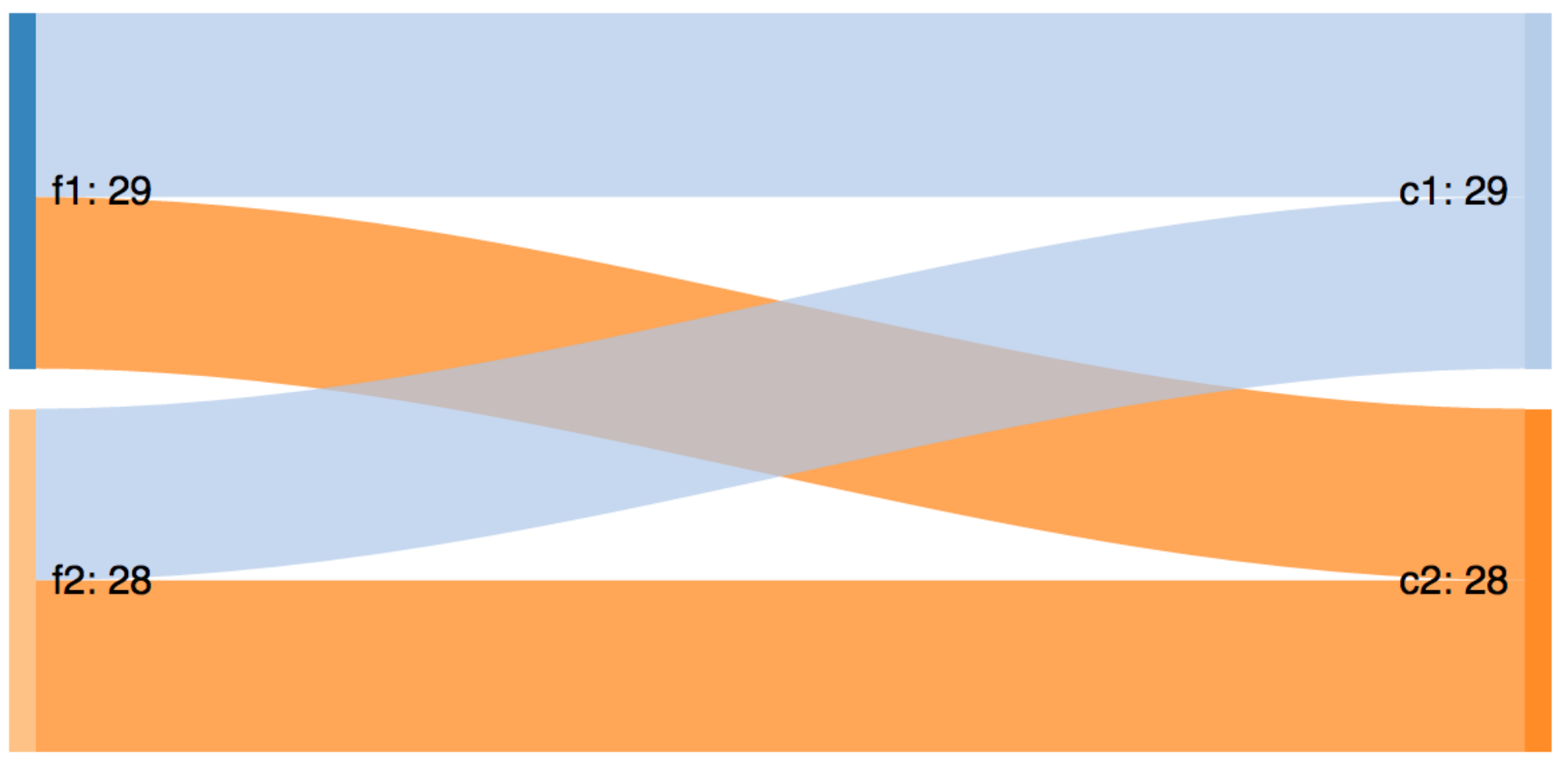}}        
    \end{subfigure}
	~    
    \begin{subfigure}[Sankey diagram of a categorical feature carrying strong signal for a classification task]
        {\label{fig:sankey_rel}\includegraphics[width=0.45\linewidth]{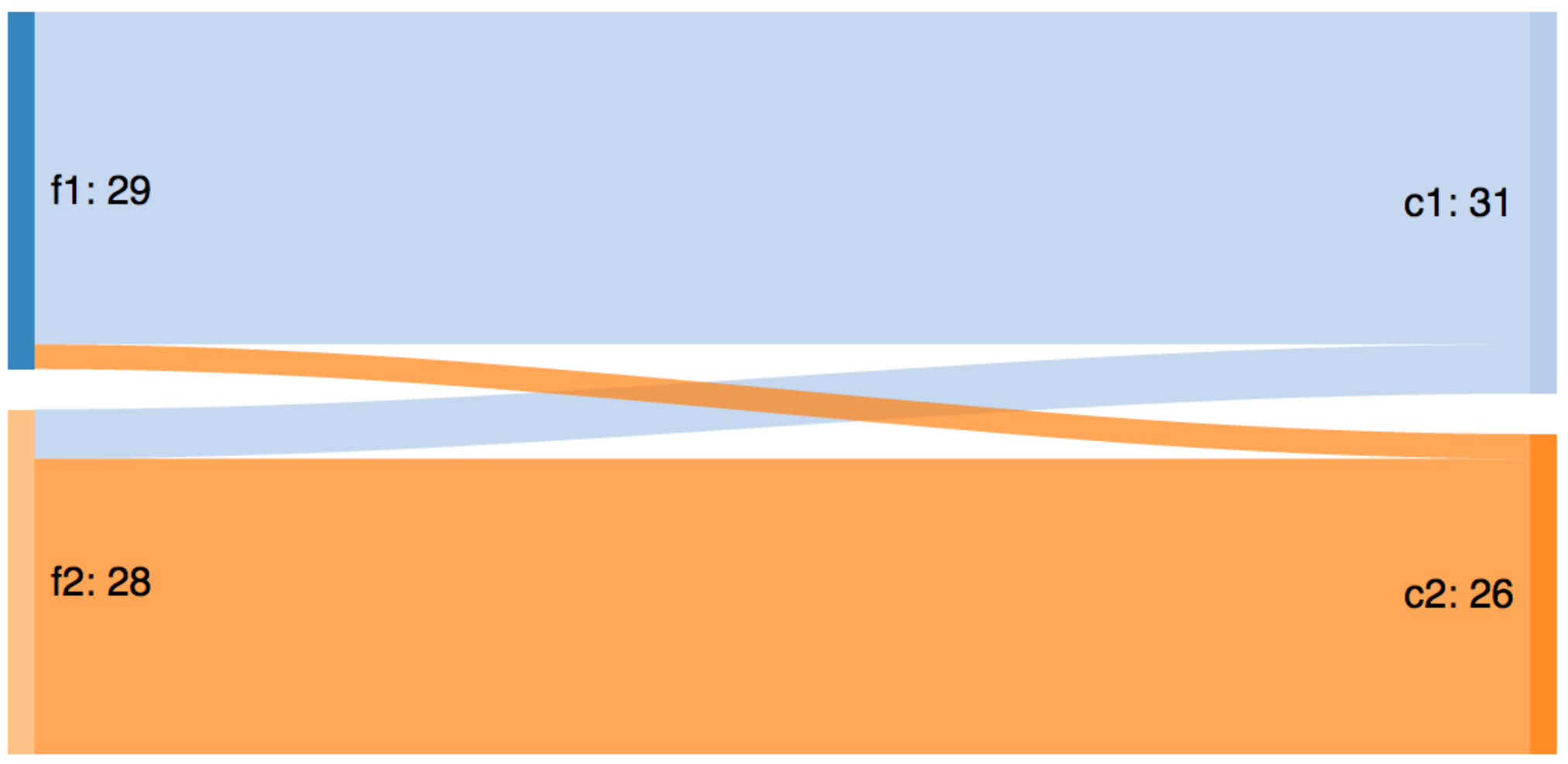}}        
    \end{subfigure}
    \caption{Use of Sankey diagram to explore feature-label relationship}
    \label{fig:example_1}
\end{figure}

\textbf{Categorical Features:} Categorical features assume a number of discrete values. Let us begin with the common scenario of involving the feature and the label spaces in a two-class classification problem. Let us denote the class labels by $y_1$ and $y_2$. $f_1$ and $f_2$ are two possible values of a categorical feature $F$. How the  $f_i$s are related to $c_j$s for all $i,j$ indicates if $F$ will be an important contributor to a classification model for this problem. For example, If $n(f_1 \rightarrow y_1) \sim n(f_1 \rightarrow y_2)$ and $n(f_2 \rightarrow y_1) \sim n(f_2 \rightarrow y_2)$, then we can infer that $F$ does not contain a strong signal to distinguish between the two classes. Figure~\ref{fig:sankey_nr} shows how Sankey diagram can be used to capture this. An example contrary to this is shown in Figure~\ref{fig:sankey_rel} where the feature is potentially useful for classification.

\begin{figure}[ht]
    \centering
    \begin{subfigure}[Conditional label distribution given director name ``James Cameron"]
        {\label{fig:director1}\includegraphics[width=0.4\linewidth]{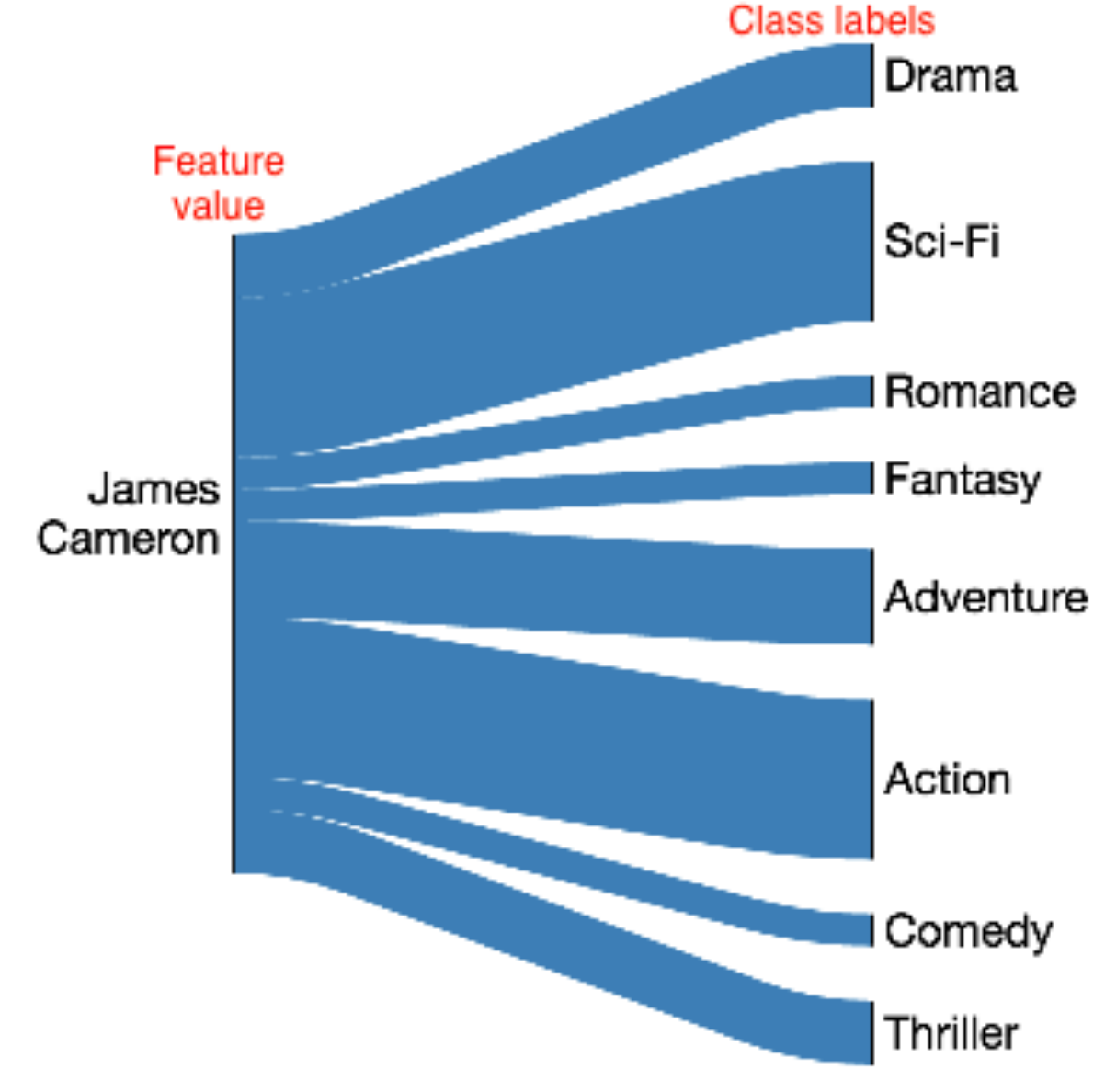}}
    \end{subfigure}
    ~ %add desired spacing between images, e. g. ~, \quad, \qquad, \hfill etc. 
    \begin{subfigure}[Conditional label distribution given director name ``Mira Nair""]
        { \label{fig:director2}\includegraphics[width=0.4\linewidth]{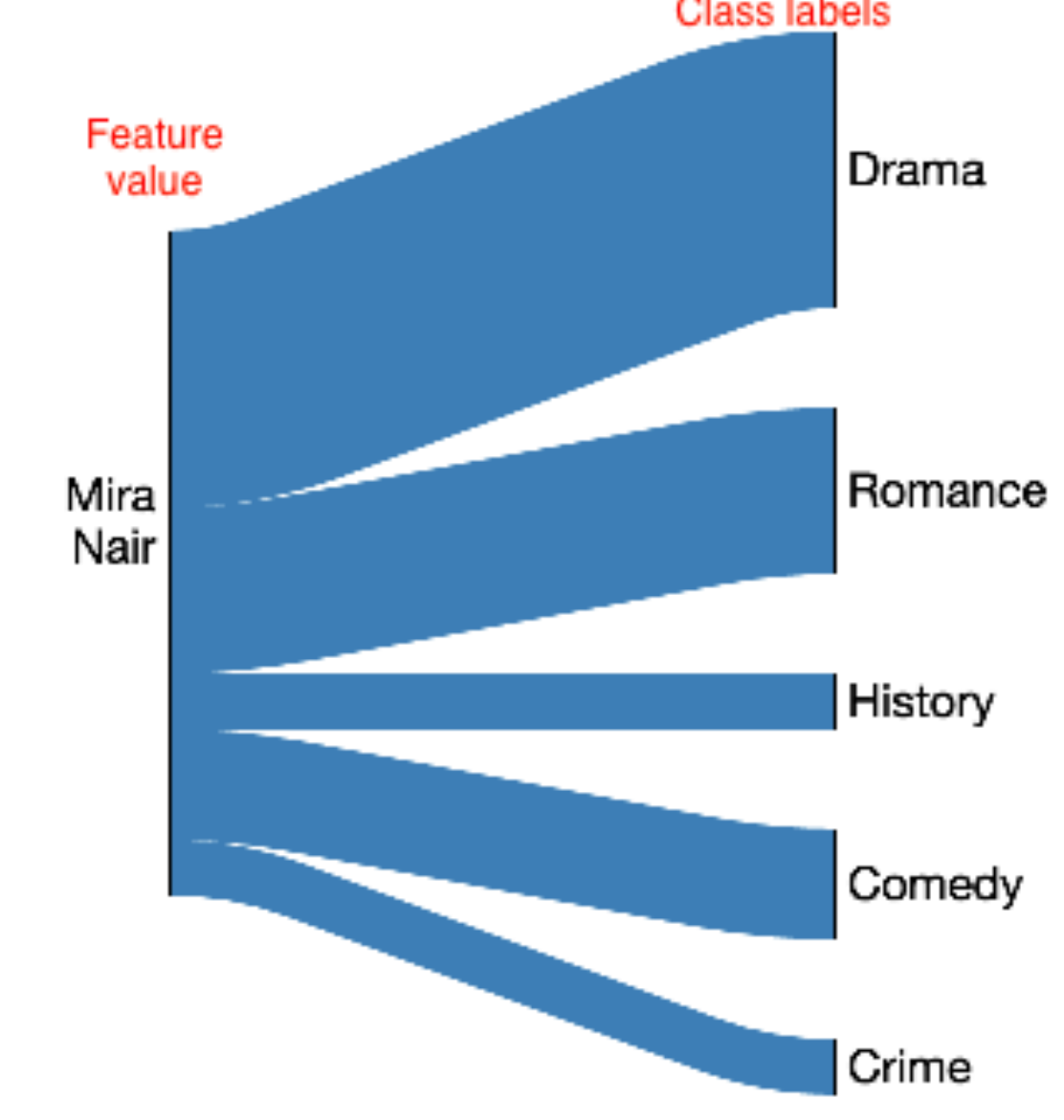}}
    \end{subfigure}
    \begin{subfigure}[Conditional label distribution for color movies]
        { \label{fig:color}\includegraphics[width=0.45\linewidth]{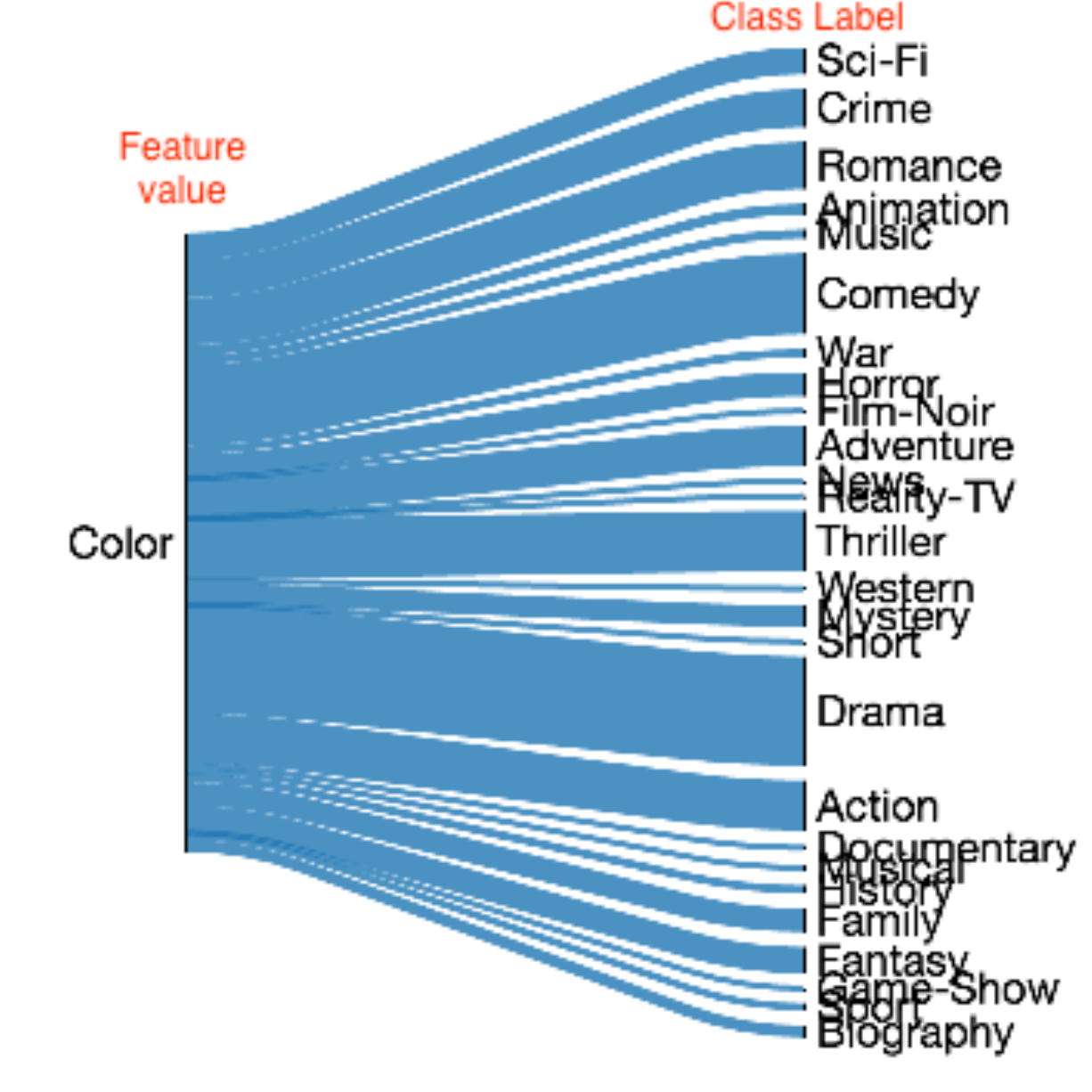}}       
    \end{subfigure}
    ~
    \begin{subfigure}[Conditional label distribution for B\&W movies]
        {\label{fig:bw}\includegraphics[width=0.45\linewidth]{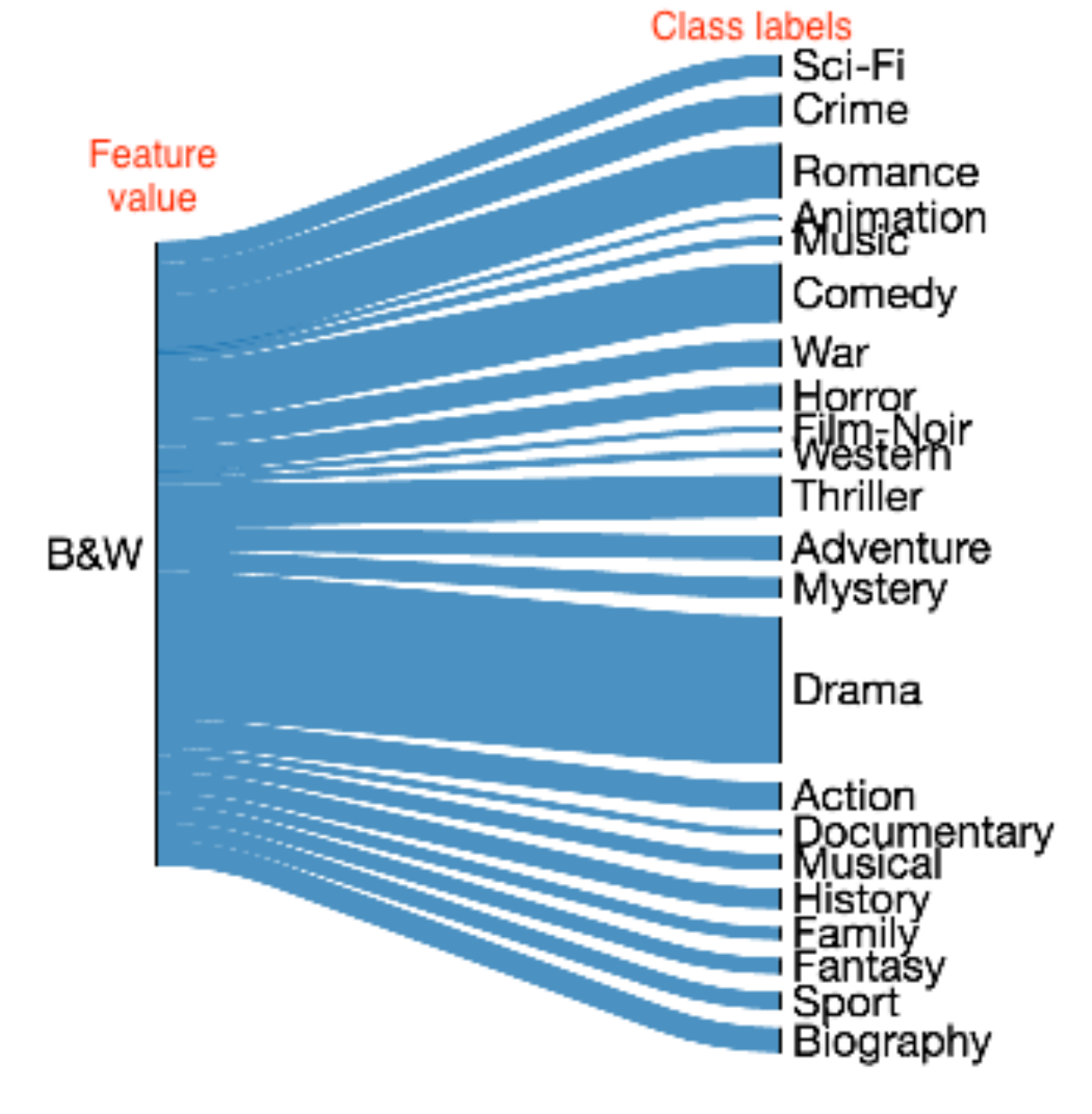}}        
    \end{subfigure}
    \caption{Use of Sankey diagram to demonstrate relative feature importance for a movie genre classification problem using IMDB dataset. \textbf{Top Row:} The distribution of genres (labels) vary significantly for different values of ``Director'' - a categorical feature. \textbf{Bottom Row:} On the contrary, the label distributions are very similar for color and B\&W - two values of ``Color''.}
    \label{fig:use_case_1}
\end{figure}

We propose to use Sankey diagrams to study the distribution of the labels for one or more values of a categorical feature. In this particular section, we deviate from our running example of product categorization dataset and use a publicly available dataset related to the publicly available movie genre classification problem~\footnote{https://www.kaggle.com/orgesleka/imdbmovies}. We study two potential features: ``Director" and ``Color" (takes two values: ``Color" or ``B\&W") are two features. Sankey diagrams in Figures~\ref{fig:director1} and ~\ref{fig:director2}  clearly highlight that the distribution of genre vary considerably from director to director, which makes it a useful feature for classifying genre. On the other feature, Figures~\ref{fig:color} and ~\ref{fig:bw} show that the distribution of genre remains largely unchanged regardless of the value of ``Color''.

\textbf{Text Features:} Many classification problems, including our product categorization use case, is largely driven by text data such as product name and description. Text data is usually converted into appropriate numeric or categorical features using suitable methods such as tf-idf or word2vec. After that, the relevance distributions of prominent keywords across categories can be visually investigated.
\subsection{Visual Analysis of Results}
\label{subsec_results}
\subsubsection{Overview of Model Evaluation} 
\label{sec_model_eval}
\begin{figure}[tb]
    \centering
     \includegraphics[width=\linewidth]{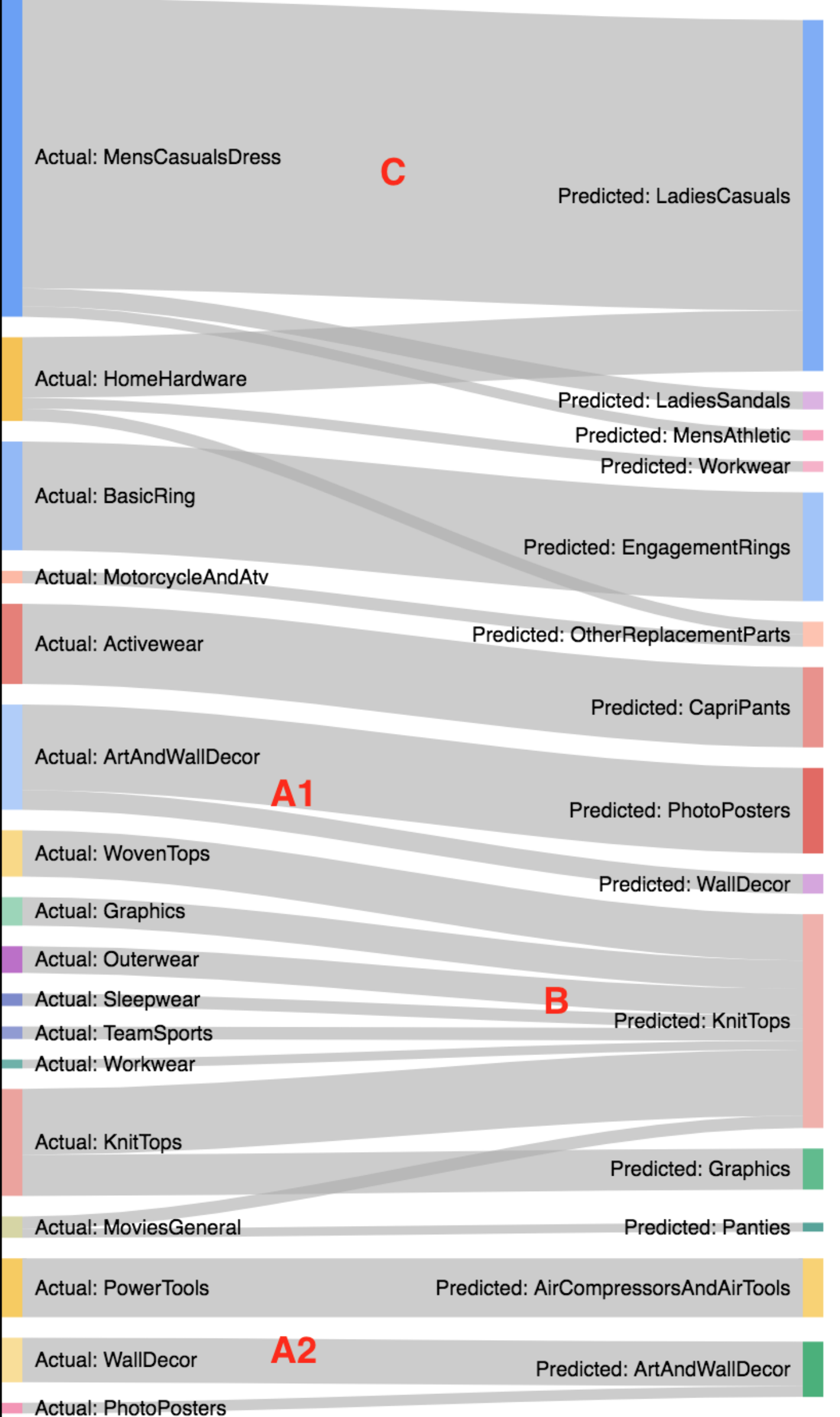}
     \caption{\textbf{Visual model diagnostics:} Visual comparison between the predicted outcomes of the classifier and the actual true labels.}
     \label{fig:model_diagnosis}
\end{figure}
Once trained, a model is evaluated on an evaluation set that consists of a relatively small yet representative set of items with known labels. The known labels that serve as ground truth in the evaluation are collected using human experts or crowd-sourcing. The model prediction for each item in the evaluation set is compared against the ground truth. The cost of each mistake is accumulated based on a distribution of importance of the items in the evaluation set. Details of the evaluation process is beyond the scope of this paper and can be found here~\footnote{https://www.youtube.com/watch?v=wkky7k4scbQ}. 

Typically, a model evaluation produces accuracy (or similar metrics) and sample size per category (Table~\ref{tbl:accu_report}). A list of mis-classified products per category is also generated, allowing fine-grained analysis (Table~\ref{tbl:err_report}).
\begin {table}[ht]
	\caption {A Segment of the Accuracy Report of a Model}
	\label{tbl:accu_report}
	\begin{center}
	\begin{tabular}{ |c|c|c| } 
	 	\hline
       Category & Sample Size & Accuracy    \\ \hline
       POP EASY LISTENING & 5 & 0.82   \\ 
       POP LOUNGE & 8 & 0.92   \\
       POP ADULT CONTEMPORARY  & 3 & 0.96   \\ 
       ... & ... & ...   \\ \hline
	\end{tabular}
	\end{center}
\end{table}
\begin{table}[ht]
	\caption{A Segment of the Misclassification Report of a Model}
	\label{tbl:err_report}
	\begin{center} 
	 \begin{tabular}{ |c|c|c| } 
     	\hline
        ID & True Category & Predicted Category    \\ \hline
		A1  & Workwear   & Sleepwear   \\
		B23  & Workwear & Men's Jumpsuit   \\
		C98  & Movie & TV Show   \\ 
        ... & ... & ...   \\ \hline
  	\end{tabular}
  	\end{center}
\end{table}
\subsubsection{Visual Error Analysis} 
\label{sec_err_report}
The class-level accuracy is relatively straightforward to visualize. A node-link visualization color-coded with class accuracies is used to represent the performance of the classifier on the different classes. 

In this paper, we emphasize on visualizing the misclassification report because it has the potential to guide the human analyst to the categories where the model consistently performs poorly. Repeated mis-classifications of similar items often indicates either a problem with the category definition or with the data. 
\begin{figure*}[tb]
    \centering
	\includegraphics[width=0.65\linewidth]{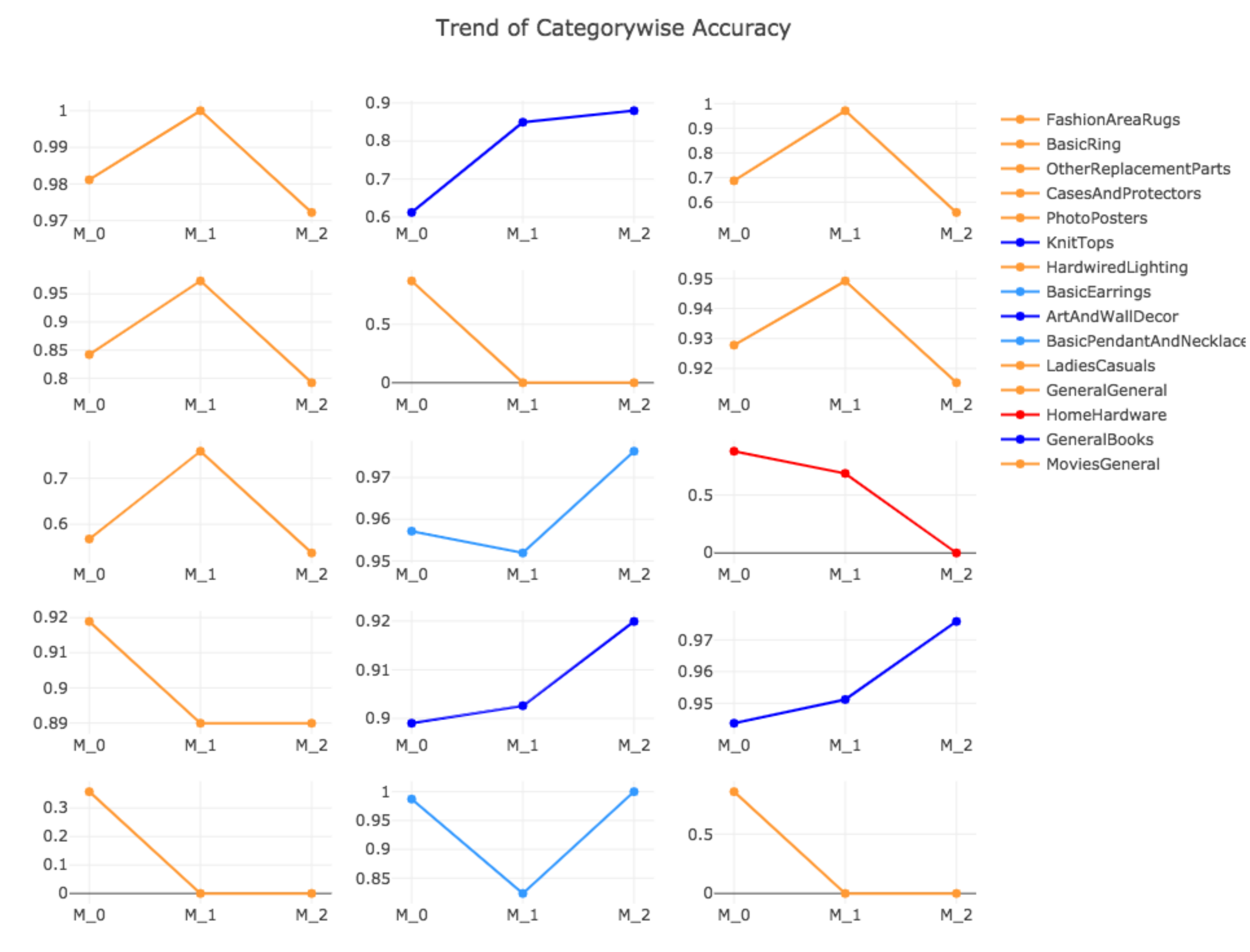}
    \label{fig:accu_trend}
    \caption{Visualization of trend of class-level accuracy over three models $M_0$, $M_1$, $M_2$ trained sequentially. Blue denotes strictly increasing trend, red denotes strictly decreasing trend. Light blue indicates an overall increasing or stable trend. Orange indicates overall decreasing or stable trend.}
\end{figure*}
\begin{figure*}[htb]
    \centering
    \begin{subfigure}[Items under KnitTops in $M_0$ are distributed among three different categories by $M_1$, suggesting an improvement. A part of PassengerTires are moved to a more generic class called OtherReplacementParts by $M_2$.]
    {\label{fig:multi_model_1}\includegraphics[width=0.4\linewidth]{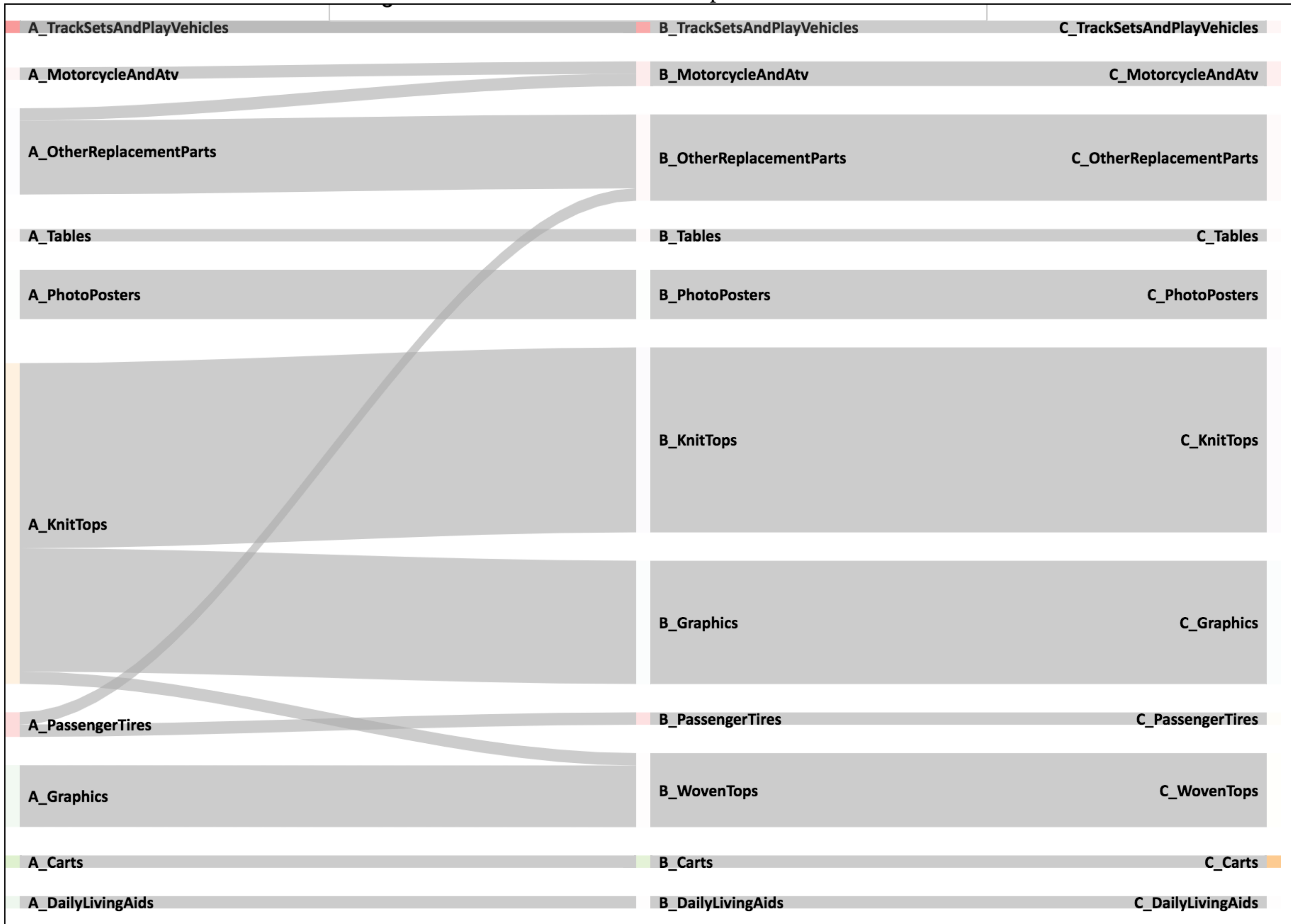}}
    \end{subfigure}    
    \begin{subfigure}[A subset of GeneralCrafts items moved to ArtAndWallDecor while a part of ArtAndWallDecor moved to SportsMemorabilia. A clear split of MensBoots into two classes is also visible between $M_0$ and $M_1$.]
        {\label{fig:multi_model_2}\includegraphics[width=0.41\linewidth]{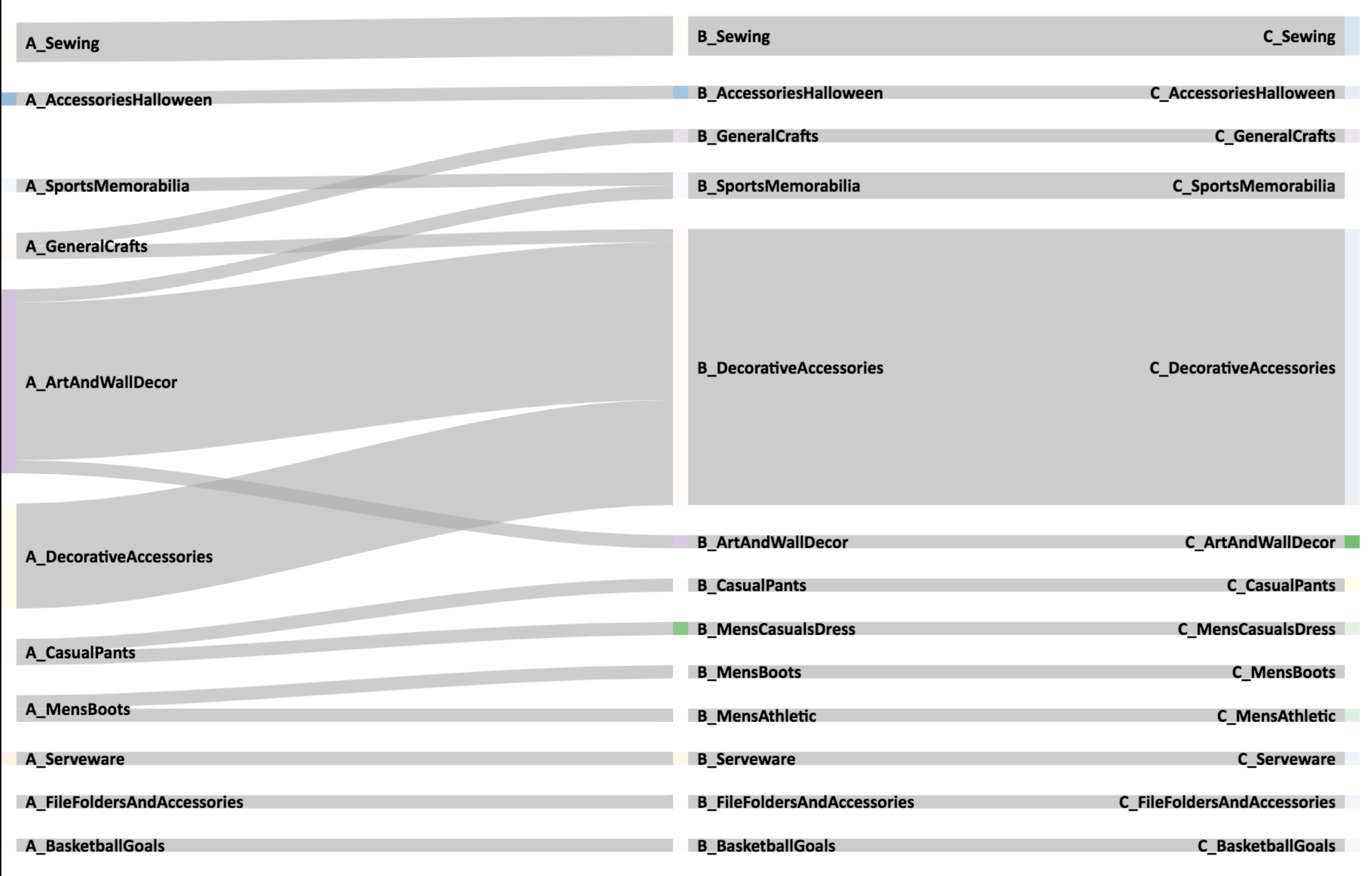}}        
    \end{subfigure}
    \caption{Visualization of potentially important prediction changes from one model to another.}\label{fig:multi_models}
\end{figure*}

We propose to draw flow diagrams between the bag of predicted labels and the bag of true labels for the mis-classified items for each category. Figure~\ref{fig:model_diagnosis} shows various interesting observations about the model performance. In may be noted that majority the items in a category called "Wall Decor" are consistently misclassified to another called "Art and Wall Decor" (region A1). The diagram also reveals that the misclassification is happening in both directions - a sizable portion of "Art and Wall Decor" items are being classified into "Wall Decor" (region A2). The diagnosis for the problem is that the category definitions are too close. Another example is notable where items in various clothing categories are mis-classified into "KnitTops", indicating the "KnitTops" category is too broad (region B). Also, the classifier does not have enough information to perform finer categorization. The visualization reveals more obvious mistakes such as "MensCasualDress" being mis-categorized in "LadiesCasuals" indicates that the model is being trained on wrongly labeled data (region C).
\begin{figure*}[!tb]
    \centering
     \includegraphics[width=0.8\linewidth]{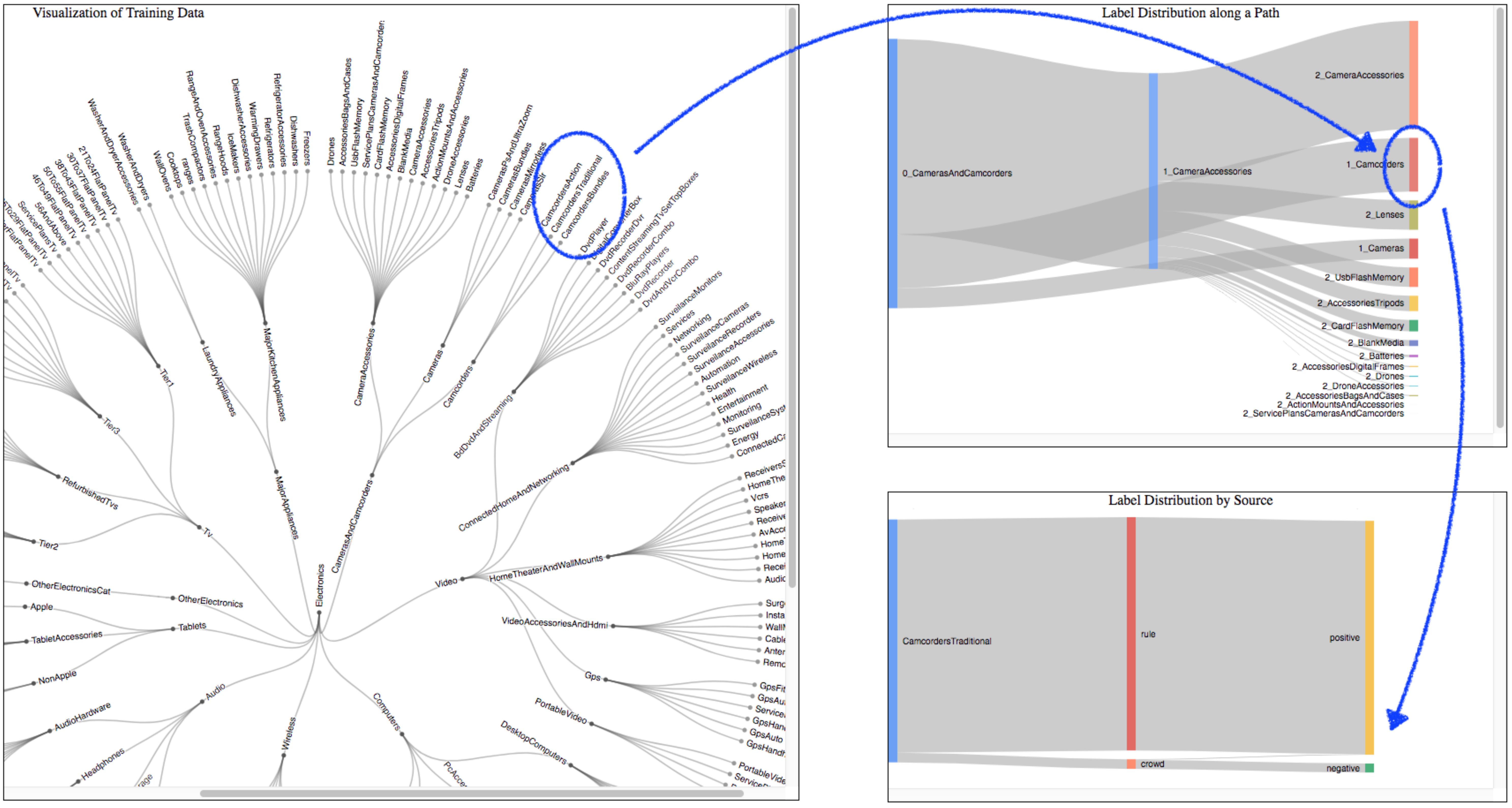}
     \caption{Web-based visual interface for exploring the training dataset. The circles with the blue arrows highlight an example flow of exploration. The radial node-link diagram (left) helps the user select a node. The distribution of label quantity of that node and its siblings is shown on top-right. The user can further explore the quality of the labels in terms of source for each category using the bottom-right view.}
     \label{fig_visp_1}
\end{figure*}
\subsection{Comparative Analysis of Multiple Models}
\label{subsec_results_multi_models}
Machine learning models are often improved incrementally. It is a common practice to train multiple models simultaneously, or over a period of time for the same problem by varying one or more of the following: training data, features, hyper-parameters, and algorithm. In the context of e-commerce, product data come with two strong textual sources of information: product title and product description. Product image is another rich source that can be leveraged. Hence, it is worth experimenting if product categorization improves if both title and description are used as opposed to only title. Transitioning to a deep learning approach is another step that requires extensive comparison with the current model.

When comparing two or more models, the overall accuracy number does not provide enough information for comparison, especially when the class structure is large. Models need to be analyzed and compared at class level and at instance level. Visual methods are ideal for such detailed analysis.

Our case study is based on three models $M_0$, $M_1$, and $M_2$ where $M_0$ and $M_1$ differ in the feature set, $M_2$ uses a deep neural network as opposed to the first two that are logistic regression based.

The first step is to compare class-level accuracies. We choose to use line charts for each class to highlight the trend of accuracy change (Figure~\ref{fig:accu_trend}). Accuracy trends of multiple classes are observed together by presenting these line charts together as small multiple plots. The line charts are color coded based on the type of the trend to immediately draw attention to classes that require human attention. An example in Figure~\ref{fig:accu_trend} is "HomeHardware" that is shown in red because of its strictly decreasing accuracy.  

The accuracy numbers at class level can be misleading. For example, the accuracy of a class can stay more or less same while the items misclassified by the model for that class can change significantly. Hence, with a new model in hand, it is important to study the major changes in predicted labels across various classes. A model that brings forward drastic changes to large groups of items often needs more scrutiny before deployment. However, given that the evaluation set often contains hundreds of items per category, it is impossible for a human analyst to go over each individual mis-prediction or change of prediction. We propose to use a visualization that aims to guide the analyst to the potentially interesting subsets of the evaluation set.

Figure~\ref{fig:multi_model_1} and Figure~\ref{fig:multi_model_2} shows that Sankey diagrams can be concatenated to capture if the predictions of a large bag of items in the evaluation set change as we switch from one model to another. Items from models $M_1$, $M_2$, and $M_3$ are plotted from left to right. If $M_2$ is able to assign finer sub-categories for to certain items that were in a broader category in $M_0$, the diagram is able to highlight that as a splitting flow. The split of "KnitTops" in Figure~\ref{fig:multi_model_1} into three categories is one such example. On the other hand, Figure~\ref{fig:multi_model_2} shows that a large group of items classified as "ArtAndWallDecor" by $M_1$ joined a broader category in $M_2$. Such changes often demand further investigation of individual instances by a human expert. 
\section{Application in a Machine Learning System}
\label{sec_application}
This section outlines how the proposed techniques are currently in use in accordance with machine learning algorithms, and how it can be extended further.
\subsection{Implementation Details}
\label{sec:impl}
At present, the visualization modules are implemented in Javascript using Google Charts API\footnote{https://developers.google.com/chart/} (for Sankey diagrams), d3.js\footnote{https://d3js.org/} (for circular node-link layout), and Plotly\footnote{https://plot.ly/} (for small multiple plots). The data scientists often use Python with Jupyter for rapid prototyping. The JS-based visualization modules run based on data produced by the Python experiments and present the visualizations on stand-alone web-based interfaces.  

To enable the data scientists to iterate between model building and visualizing even faster, a Python version of Plotly, along with a widget for Sankey diagrams~\footnote{https://github.com/ricklupton/ipysankeywidget}, is used to integrate the visualizations directly within the Jupyter workflow. 
\subsection{Outline of Interactive System}
\label{sec:vis_components}
The system currently has three modules: one for the exploring the training data, one for exploring the classifier performance, and one for comparing multiple classifiers.

The interface for training data exploration consists of three views - a node-link style network visualization of the training data with radial layout, a Sankey diagram visualizer for a group of nodes, and another Sankey diagram visualizer for the label distribution of a single node. Figure~\ref{fig_visp_1} shows an example flow of exploration using these views. In this case, the user first explores that there are relatively large number of labels for "Camcorders" as opposed to some other categories that have very few such as "Batteries" and "Drones". However, further exploration with the bottom right view reveals that most of the labels for "Camcorders Traditional" came from rules and none came from trained experts. This is a clear signal for requesting more labels for this category. 

This design suits our application where the product catalog is a large hierarchy, and it should generalize well to most classification problems that are inherently tied to a hierarchical class structure. The node-link visualization serves as the main overview driving the system. Class-specific properties such as sample size and classification accuracy can be encoded as node color, size etc so that the user can make informed selection of nodes.

The interface for exploring the model results contains two views: a node-link visualization for showing the class-wise accuracies, and a Sankey diagram highlighting the relationship between predictions and ground truth labels.

The interface for exploring multiple models contains two views: a collection of trend charts arranged in a grid layout where each chart reflects the accuracy trend of a class. An accompanying Sankey diagram presents the major prediction changes.

\section{Conclusion and Future Work}
\label{sec_conc}
In this paper, we present a set of techniques to enhance machine learning driven classification systems with visualization and interactive data analysis. The visualizations are designed around a central theme of understanding the flow of information across different entities such as training set, features, and results. In particular, we explore the potential of using flow diagrams, namely Sankey diagram, to capture the flow of information in a machine learning system. Other types of flow diagrams such as chord diagram can be applied as well depending on the application. The examples and datasets in the paper are derived from a real application: large-scale product categorization in e-commerce. The proposed technique can benefit similar hierarchical classification systems in other domains as well. Also, the proposed technique is not tied to any particular machine learning algorithm. 

We are in the process of improving our system in many ways. We plan to integrate the web-based visual interfaces into a central system. At present, most of the data selection operations are driven from the Python scripts running at the back end. We plan to add more interactive capability to the front end. We also plan to allow the user to trigger a data filtering task or a model training task directly from the interface. We have received positive feedback about the system from various corners. However, we plan to conduct more formal experiments and user studies to quantify the benefit of such a system. We also plan to extend this system for understanding other machine learning driven systems, especially the deep learning based systems. 

\bibliographystyle{abbrv-doi}
\bibliography{template}


%\begin{biography}
%Abon Chaudhuri develops machine learning driven solutions for large-scale
%classification and prediction problems in a Data Science Team at Walmart
%Labs, Silicon Valley. Earlier, he worked on analyzing and visualizing
%nano-scale imaging data at Intel Corporation. He has experience of
%visualizing scientific data produced by supercomputers at Argonne and Oak
%Ridge National Laboratory. Abon received his PhD in Computer Science and
%Engineering from The Ohio State University in 2013. His doctoral research
%focused on feature-based summaries of big data to enable
%visualization and query-driven analytics. Abon has authored several
%research publications that earned him three best paper/poster awards and
%a visualization award at top-tier IEEE conferences.
%\end{biography}

\end{document}